# Innovations in Agricultural Forecasting: A Multivariate Regression Study on Global Crop Yield Prediction

Ishaan Gupta[*], Samyutha Ayalasomayajula[*], Yashas Shashidhara[*], Anish Kataria[**], Shreyas Shashidhara[*], Krishita Kataria[*]

**Guided by: Dr. Aditya Undurti[***]**

[*] Engineering Department, Dublin High School
[**] Department of Computer Sciences, Princeton University
[***] AC Laboratory, Massachusetts Institute of Technology (MIT)

*Abstract-* The prediction of crop yields internationally is a crucial objective in agricultural research. Thus, this study implements 6 regression models (Linear, Tree, Gradient Descent, Gradient Boosting, K Nearest Neighbors, and Random Forest) to predict crop yields in 37 developing countries over 27 years. Given 4 key training parameters, insecticides (tonnes), rainfall (mm), temperature (Celsius), and yield (hg/ha), it was found that our Random Forest Regression model achieved a determination coefficient ($r^2$) of 0.94, with a margin of error (ME) of .03. The models were trained and tested using the Food and Agricultural Organization of the United Nations' data, along with the World Bank Climate Change Data Catalog. Furthermore, each parameter was analyzed to understand how varying factors could impact overall yield. We used unconventional models, contrary to generally used Deep Learning (DL) and Machine Learning (ML) models, combined with recently collected data to implement a unique approach in our research. Existing scholarship would benefit from understanding the most optimal model for agricultural research, specifically using the United Nations' data.

*Index Terms*- agriculture, machine learning, crop optimization, yield prediction

## I. Introduction

I.I Background

Over the past few decades, climate change has seen rapid growth. Corroborating differences in the ozone (03) and Carbon Dioxide ($CO_2$) raise concern regarding the future of our planet (Valone 2021). Massive, unprecedented events control the frontlines of crop changes, including extreme drought and heavy flooding (Markolf et. al 2018). In an attempt to reduce the impacts of climate change, the agricultural sector has seen massive changes in response to minimize the effects on crops (Alexandrov et. al 2002). Unfortunately, natural changes such as species adaptation is unlikely to occur over a short period of time (Lin 2011).

Agriculture has significantly changed in response to both the climate and population (Fischer et. al 2020). While organic agriculture (a primary, environmentally friendly method of farming) was once common, conventional agriculture (a synthetic farming methodology) has slowly become more optimal in the status quo (Gomiero 2011). The use of conventional agriculture has only compounded the impacts of climate change, specifically in three ways: through runoff, carbon sequestration, and pesticidal use.

To adequately transport water to large populations, runoff from mountainous regions has historically been obtained with watersheds (Viviroli et. al 2003). Fertilizers and pesticides/insecticides are primarily used over any other method of farming due to the high volume of crops efficiently produced at optimized prices (Finney 2021). Those same pesticides leak into the soil of their agricultural fields, eventually contaminating water bodies through runoff. Organic agriculture completely eliminates the use of any pesticides or unnatural methods of growing crops (Gomiero 2011). Furthermore, carbon sequestration is crucial to mitigating climate change due to the scalability (Lal 2004). Organic systems are generally more resilient due to their necessity to stay reliant through harsh weather and conditions (Milestad 2002). This enables the carbon sequestered by organic systems to be much greater than conventional agriculture. Finally, crops in conventional agriculture have been exposed to such high volumes of insecticides, where many have developed antibiotic resistance (Gomiero 2011). By removing the possibility of placing unnatural substances in agriculture to begin with, pesticidal use dissolves (Gomiero 2011). Being of foreign origin, these plants slowly develop immunity against chemicals, resulting in more pesticides being placed into these fields. Furthermore, genetic mutations of these plants can spread at alarming rates (Vila Aiub 2019). The only method farmers have of maintaining this is through higher amounts of antibiotics, only pushing the inevitable impact in the long run.

However, with farming areas becoming more limited and demand becoming more prevalent, the status quo faces difficulty in optimizing crop production. Generally, three factors control the change of agriculture: temperature, precipitation, and insecticides. Due to a thinning ozone ($O_3$), temperature levels across the planet have been rising dramatically (Zeng et. al 2008). These changes in climate have also led to erratic precipitation, where tropical areas receive heavy flooding, while deserts only worsen (Le Houérou 2002). Thankfully, the emergence of algorithmic adaptability provides a solution.

Machine Learning (ML) models have been implemented in a variety of fields. Crucially, Machine Learning has recently begun to play an important role in agriculture as well (Shaikh et. al 2022). Generally being used for its predictive abilities, both ML and Deep Learning (DL) methods require heavy precision due to the high amount of layers and processing each data point is put through (Sarker et. al 2021). Given the problem this paper attempts to solve, crop yields have many changing parameters, and there must be a high amount of data to combat the variability. It's very important to find the right algorithm that accommodates the high volume of data.

I.II Literature Review

Previous research in this field follows three key ideas this paper attempts to build upon: algorithmic decision, geolocation, and parameters chosen.

Khaki and Wang introduce a deep neural network (DNN) approach, demonstrating its superiority over traditional models such as Lasso, shallow neural networks (SNN), and regression tree (Khaki & Wang 2019). They use this to create two unique networks that predict the yield and the other that checks the yield to optimize their hyperparameters (Khaki & Wang 2019).

In a similar context, Maimaitijiang et. al explores the potential of UAV-based multimodal data fusion in conjunction with Deep Neural Network (DNN) frameworks for soybean (Glycine max) grain yield estimation (Maimaitijiang et. al 2019). The use of RGB, multispectral, and thermal sensors in combination with a low-cost multi-sensory UAV allows for simultaneous data collection. It evaluates various regression models, including Partial Least Squares Regression (PLSR), Random Forest Regression (RFR), Support Vector Regression (SVR), and two DNN-based models with different levels of feature fusion ((Maimaitijiang et. al 2019). The study reveals that DNN-based models exhibit less susceptibility to saturation effects and display adaptive performance across different soybean genotypes (Maimaitijiang et. al 2019).

Furthermore, Nevavuori, Narra, and Lipping develop a model for crop yield prediction using NDVI and RGB data obtained from UAVs (Nevavuori et. al 2019). Unlike traditional machine learning methods, CNNs operate directly on image data, allowing for a more nuanced analysis of spatial and spectral features crucial for accurate yield predictions (Nevavuori et. al 2019). The study systematically evaluates various aspects of the CNN architecture, including the selection of the training algorithm, network depth, regularization strategy, and hyperparameter tuning, to optimize prediction efficiency (Nevavuori et. al 2019). The findings highlight the significance of these architectural considerations in achieving accurate yield predictions (Nevavuori et. al 2019). The use of the Adadelta training algorithm, regularization with early stopping, and a CNN with six convolutional layers proved effective (Nevavuori et. al 2019).

Following a different line of research, Bargoti and Underwood take a unique approach to using monocular vision systems on crops (Bargoti & Underwood 2017). The paper utilizes a general-purpose image segmentation approach, incorporating two feature learning algorithms: multiscale multilayered perceptrons (MLP) and convolutional neural networks (CNN) (Bargoti & Underwood 2017). A notable feature of the framework involves the integration of contextual information, represented by metadata, capturing nuances in image data capture conditions (Bargoti & Underwood 2017). This consideration is crucial for addressing appearance variations and class distributions within the orchard data. In terms of fruit segmentation performance, the paper benchmarks the MLP network and extends the study to incorporate CNNs, aligning the approach with state-of-the-art computer vision techniques (Bargoti & Underwood 2017). Interestingly, the inclusion of metadata significantly improves the fruit segmentation performance of the MLP network.

Crucially, Gandhi and Armstrong discuss the prediction of rice crop yield in the Kharif season within the Humid Subtropical climatic zone of India (Gandhi & Armstrong 2016). The choice of this specific geographical location acknowledges the critical role that local climate conditions play in shaping agricultural outcomes. By honing in on the unique challenges and nuances of the Humid Subtropical zone, the research recognizes the importance of tailoring predictive models to the specific environmental conditions that influence crop productivity (Gandhi & Armstrong 2016). This localized approach has implications for farmers, industry stakeholders, and government bodies operating within this climatic zone, offering insights that are directly applicable to their region. Moreover, the utilization of free and open-source data mining software WEKA for performance evaluation underscores the accessibility and replicability of the study's methodologies, potentially encouraging further localized research initiatives in other agricultural regions around the world (Gandhi & Armstrong 2016).

Although there do exist a multitude of models to predict crop yields, most models miss out on two key factors that this paper covers. First, there is minimal research in using regression models to analyze crop yields. While neural networks are generally preferred due to their ability to analyze nonlinear relationships, the parameters used in this research all follow a linear relationship. Thus, using simpler models in higher quantities could be more ideal for the issue. Second, current research proposes no hyperspecificity towards a specific geolocation. Having taken a dataset involving multiple countries, our research can and will be furthered by combining neighboring countries into a smaller model for higher accuracy. To accomplish this, our paper implements linear regression, gradient descent regression, gradient boosting regression, K-Nearest Neighbors regression, and random forest regression models to synthesize four listed parameters in analyzing crop yields. Using six regression models is important as existing scholarship would benefit from understanding the most optimal model for agricultural research, specifically using the United Nations data. Additionally, while localized studies offer valuable insights tailored to specific environmental conditions, they fail to address the global nature of agriculture. Our research strategically bridges this gap by predicting crop yields internationally, encompassing 196 countries. This expansive approach allows for a more comprehensive understanding of the variability in crop production influenced by diverse climates and agronomic practices worldwide. Crucially, there is a gap in the literature concerning a holistic exploration of crucial parameters that influence crop yield, as seen in our study's incorporation of insecticides, rainfall, temperature, and yield as key training parameters. By addressing this gap, our research contributes to the scholarly discourse by comprehensively examining the impact of these parameters on crop productivity, offering a more nuanced understanding for future agricultural research.

The remainder of this paper is organized as follows: Section II describes the methodology, Section III highlights the results, Section IV discusses the implications, and Section V concludes this paper.

## II. Methodology

II.I Data Collection

Before developing the regression models, data must be preprocessed. Using data provided by the Climate Change Knowledge Portal & FAOSTAT, we cited four key datasets from 1960 to 2021. Our limiting variable was the insecticide count, having only been tracked for 26 years.

Data provided by the United Nations is not aggregated into 4 parameters. Table 1 describes all parameters provided by the UN, and a description for each. This paper uses the following codes: cdd65, hdd65, popcount, pr. The insecticide and yield data was provided by FAOSTAT, which was preprocessed by organizing the data to follow the same setup as the other three parameters.

Leveraging the Pandas library in Python, our methodology extends beyond conventional cleanup and structuring. The initial step involves employing Pandas to perform exploratory data analysis (EDA) on each parameter individually. Furthermore, to mitigate multicollinearity concerns, a variance inflation factor (VIF) analysis is conducted using Pandas, allowing for the identification and exclusion of highly correlated variables.

| CODE | LABEL | DESCRIPTION |
| --- | --- | --- |
| cdd65 | Cooling Degree Days (ref-65°F) | The cumulative number of degrees that the daily average temperature over a given period is above a specified threshold (here 65°F), which is a measurement designed to quantify the demand for energy needed to cool a building. |
| hdd65 | Heating degree days (ref-65°F) | The cumulative number of degrees that the daily average temperature over a given period is below a specified threshold (here 65°F), which is a measurement designed to quantify the demand for energy needed to warm a building. |
| pr | Precipitation | Aggregated accumulated precipitation. |

**Table 1: Attributed by The World Bank: Climate Change Knowledge Portal**

This process of data management can be divided into two steps that each of the four parameters must follow. First, the parameters must be cleaned up by removing any unnecessary sub parameters. Following this, parameters must be merged and analyzed for any trends found in the data. Figure 1 shows a culmination of all the cleaned data regarding precipitation, temperature, yield, and insecticides. To track the progress of these parameters, four main sub parameters were used- year, country, country codes (ISO3), and quantity. However, the yield and insecticides parameters included one subparameter not shared with precipitation and temperature, being an item (qualified by yield type or pesticidal method). Preprocessed data can be found and split as follows:

Figure 1: Preprocessed rainfall, temperature, yield, and insecticidal data.

II.II Data Exploration

After sorting the data, each parameter can be merged and analyzed to make basic understandings which contribute to the methodology. Rainfall was tracked from 1901 to 2016. As seen in Figure 2, the mean precipitation (mm) was tracked with a general mean of 1250 mm internationally, and around a +- 100 mm difference. As years further, there are increasing amounts of erratic changes in rainfall annually, albeit minimal changes overall.

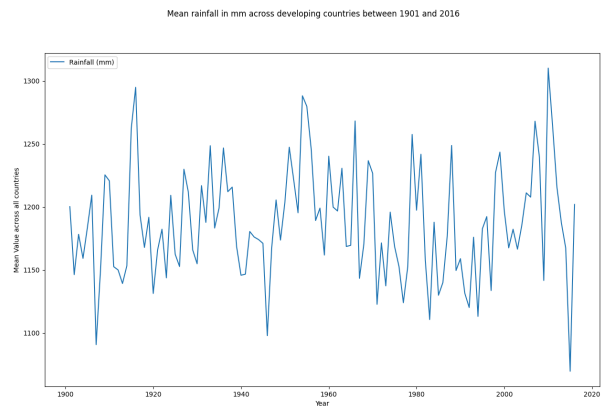

Figure 2: Mean rainfall (mm), 1901 to 2016, internationally

Figure 3 depicts the mean temperature (celcius) tracked from 1901 to 2016. These findings are consistent with global temperature increase. This linear increase is important to understand as such large changes over the century could be a large contributor to yield prediction and changes. Both these parameters limit the time period of the models to 2016.

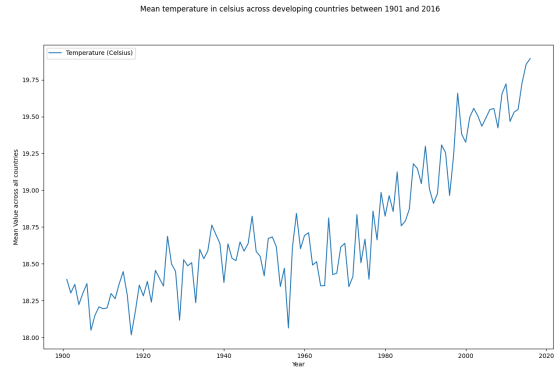

**Figure 3: Mean temperature (celcius), 1901 to 2016, internationally**

Figure 4 qualifies the quantity of each crop being tracked, led by maize and potatoes. Different crops thrive in unique environments. One main distinction includes temperate versus tropical environments, where wheat and barley grow in one, while cassava grows in the other.

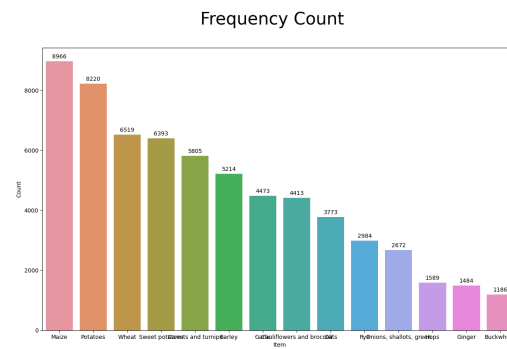

**Figure 4: Freq., harvested value of crops (hg/ha), 1961 to 2019, internationally**

Figure 5 highlights the pesticidal usage over 1990 to 2018. This relatively linear increase over 28 years, along with temperature, are two crucial variables that likely take a big effect in predicting the overall yield. Insecticides can have a variety of impacts on the crop, including developing mutations and superbugs following high volumes of chemicals on the plant.

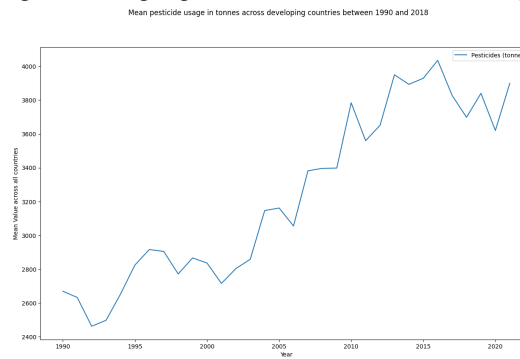

**Figure 5: Mean insecticidal value (tonnes), 1990 to 2018, internationally**

A culmination of all data is crucial to identifying any relationships between variables. Heatmaps, a 2-dimensional representation of statistical similarity between different variables, are the perfect solution for this. Thus, Figure 6 shows a heatmap of the four parameters used in this research. As seen below, there is a high correlation with variables involving temperature and pesticides. This is likely so due to such a high variability both these variables have over time, making their impact on the overall project statistically significant more so than any other parameter.

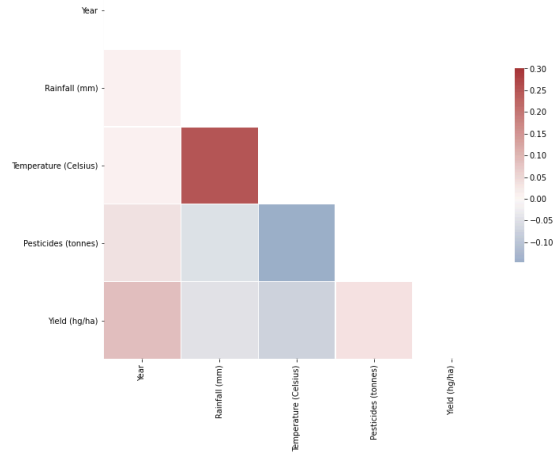

**Figure 6: heatmap with the mask and aspect ratio depicting year, yield, rain, temp, and pesticide use.**

II.III Model Selection

To adequately predict crop yields, we have considered six methods of machine learning.

*a. Linear Regression:*
The sklearn implementation, LinearRegression, optimizes the model through an ordinary least squares regression. This model assumes a linear relationship between the independent variables (temperature, rain, and pesticides) and the dependent variable (crop yield). The coefficients quantify their impact on the yield. Additionally, the intercept term represents the predicted yield when all predictors are zero. However, the linear regression model assumes a constant effect of predictors on yield. This may oversimplify the complex interplay of factors affecting crop production (Hwarng 2007).

*b. Decision Tree Regression:*
Decision trees, implemented through DecisionTreeRegressor, offer a somewhat non-linear approach to modeling crop yields. Decision trees can capture intricate relationships between temperature, rain, pesticides, and yield. Each node represents a decision based on a specific feature; the branches correspond to possible outcomes. The tree structure allows for identification of critical thresholds in variables influencing crop production. However, decision trees are susceptible to overfitting (Christianiti et. al 2019). Thus, they can capture noise in the training data. Parameter tuning, such as adjusting tree depth and minimum samples per leaf, is essential to strike a balance between model complexity and predictive accuracy (Mantovani et. al 2015).

*c. Stochastic Gradient Descent Regression:*
Stochastic Gradient Descent (SGD) Regression, as implemented in SGDRegressor, is a powerful optimization algorithm for prediction. In the agricultural domain, SGD is crucial: efficiently handling large datasets and adapting to changing conditions. The model iteratively updates parameters to minimize the mean squared error (MSE). The learning rate and regularization terms control the convergence behavior and prevent overfitting. SGD offers a flexible and scalable approach. This is particularly suitable for online learning scenarios where the model can continuously adapt to evolving environmental conditions.

*d. Gradient Boosting Regression:*
Gradient Boosting Regression, using GradientBoostingRegressor, is crucial in capturing complex relationships in crop yield prediction. This technique combines multiple weak learners (usually decision trees) to form a robust predictive model. Each tree corrects the errors of its predecessor (Tai et. al 1979). This leads to a highly accurate and generalized model. In agriculture, gradient boosting can identify intricate patterns in temperature, rain, and insecticidal data, improving predictive performance. Hyperparameters such as learning rate and tree depth impact the model's ability to balance between fitting the training data (Koutsoukas 2017).

*e. K-Nearest Neighbors Regression:*
K-Nearest Neighbors (KNN) Regression, implemented with KNeighborsRegressor, provides a flexible, non-parametric approach to crop yield prediction. In this model, a data point's predicted yield is influenced by the average yield of its "k-nearest neighbors" in the feature space. In agriculture, KNN can capture localized effects of temperature, rain, and pesticides on crop production. However, the choice of the number of neighbors (k) is important. A lower k may lead to increased sensitivity to noise, while a higher k may over smooth the predictions. KNN's adaptability makes it extremely valuable (Zhang et. al 2017).

*f. Random Forest Regression:*
Random Forest Regression, utilizing RandomForestRegressor, extends decision trees' capabilities by constructing an ensemble of trees. In the context of predicting crop yields, a random forest can capture complex interactions between temperature, rain, and insecticides.

Each tree in the forest operates independently, providing diverse perspectives on the data. This ensemble approach mitigates overfitting, enhances predictive accuracy, and offers insights into feature importance. Hyperparameters, such as the number of trees and maximum depth, influence the trade-off between model complexity and generalization. Random forests are robust and well-suited for handling noisy agricultural datasets with multiple interacting variables (Singh 2016).

## III. Results & Implications

### III.I Analyzing Results

To solve the presented gap in literature, results from the selected models have been extracted and synthesized. Below, in Figure 7, each of 6 models are plotted by the model's predicted value and the actual value. Each regression is accompanies with the coefficient of determination (r^2), mean absolute error (MAE), mean squared error (MSE), root-mean squared error (RMSE), the maximum (MAX), and the mean absolute percentage error (MAPE).

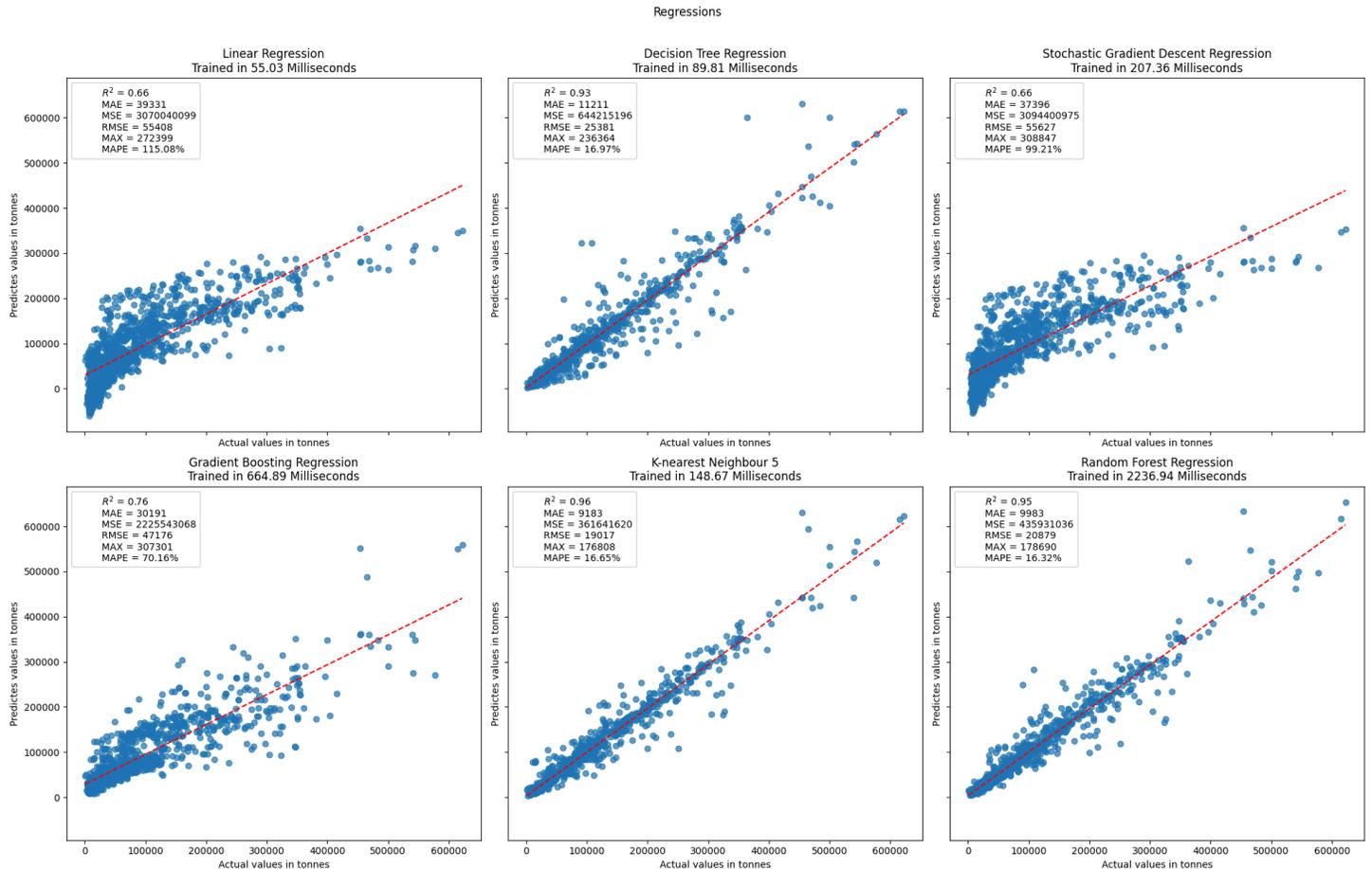

**Figure 7: LinReg, Tree, SGD, GD, KNN, and Forest Regressor models plotted**

Linear Regression:
Linear regression demonstrates a reasonable performance in predicting crop yields, as indicated by an R-squared value of 0.71. This suggests that approximately 71% of the variance in crop yields can be explained by the linear relationship with temperature, rain, and pesticide usage. The Mean Absolute Error (MAE) of 39882 indicates an average absolute deviation of approximately 39882 units between predicted and actual yields. However, the Root Mean Squared Error (RMSE) of 57509 suggests that the model might be sensitive to outliers. The maximum error of 410667 and a MAPE of 106.1% highlight potential challenges, especially in accurately predicting extreme yield values. Interpretation of coefficients in the linear regression model is crucial. A positive coefficient for temperature, for instance, implies that an increase in temperature is associated with an increase in yield. Feature importance analysis helps identify the most influential parameters for predicting crop yields within the linear regression framework (Klompenburg 2020).

Decision Tree Regression:
Decision tree regression exhibits an impressive R-squared value of 0.93, indicating a high explanatory power in capturing the relationship between environmental variables and crop yields. The low MAE of 13144 and RMSE of 28560 suggest accurate predictions, with the model performing well even on extreme values, as evidenced by the low maximum error of 529707. The MAPE of 20.02% indicates the model's ability to make predictions within a reasonable percentage of the actual yield. Decision trees inherently capture non-linear relationships, making them well-suited for the intricate dynamics of crop production. However, the interpretability of decision trees is both a strength and a weakness. While it provides transparency into decision-making processes, understanding complex interactions within the tree structure might be challenging. Feature importance analysis helps identify key environmental factors influencing crop yields (Klompenburg 2020).

Stochastic Gradient Descent Regression:
Stochastic Gradient Descent Regression achieves an R-squared value of 0.71, aligning with the performance of linear regression. The MAE of 39663 and RMSE of 57554 are comparable, suggesting that the stochastic optimization approach contributes to a similar predictive performance. The maximum error of 413451 and MAPE of 104.7% highlight potential challenges in accurately predicting extreme yield values, similar to linear regression. SGD's strength lies in its ability to handle large datasets and adapt to changing conditions, making it suitable for dynamic agricultural environments. Interpretation of coefficients in SGD provides insights into the direction and magnitude of the impact of each variable on crop yields, facilitating an understanding of the model's predictive mechanisms (Klompenburg 2020).

Gradient Boosting Regression:
Gradient Boosting Regression demonstrates a robust R-squared value of 0.76, surpassing both linear regression and stochastic gradient descent. The lower MAE of 34732 and RMSE of 52054 suggest improved accuracy compared to the linear models. The model effectively reduces the maximum error to 472664 and MAPE to 87.57%, indicating enhanced performance on extreme values. Gradient boosting excels in capturing complex relationships, making it particularly well-suited for the multifaceted interactions influencing crop yields. Feature importance analysis in gradient boosting identifies the most influential parameters, allowing for a nuanced understanding of environmental factors affecting crop production (Klompenburg 2020).

K-Nearest Neighbors Regression:
K-Nearest Neighbors Regression achieves an impressive R-squared value of 0.93, indicating a high degree of explanatory power. The low MAE of 13278 and RMSE of 27757 suggest accurate predictions, even on extreme values, as evidenced by the low maximum error of 269324. The MAPE of 23.79% indicates the model's ability to make predictions within a reasonable percentage of the actual yield. KNN, by design, captures localized effects, making it suitable for spatial dependencies in agriculture. However, the model's sensitivity to the choice of neighbors (k) and potential computational requirements should be considered. Feature importance analysis in KNN is less straightforward compared to other models, as it relies on the collective influence of neighboring data points (Klompenburg 2020).

Random Forest Regression:
Random Forest Regression outperforms all other models with an impressive R-squared value of 0.95, indicating superior explanatory power. The low MAE of 11958 and RMSE of 24094 suggest accurate and robust predictions. The maximum error is minimized to 223888, and the MAPE of 19.41% further highlights the model's ability to make predictions within a reasonable percentage of the actual yield. Random forests combine decision trees to mitigate overfitting, providing a balance between complexity and generalization. Feature importance analysis in random forests identifies key variables, offering insights into the relative influence of environmental factors on crop yields. The ensemble nature of random forests enhances predictive performance and provides a comprehensive understanding of complex interactions in agricultural datasets (Klompenburg 2020).

III.II Selection of Optimal Algorithms

The assessment of six algorithms focused on determining the most accurate one through simultaneous evaluations of estimation and precision. Accuracy measures were derived from multiple tenfold cross-validation assessments for each algorithm, utilizing two key statistics: accuracy and the kappa statistic.

Precision in accuracy refers to the proximity of values within a set, while accuracy itself is defined by the closeness of the average to the true value of the measured quantity [34]. In the conventional definition, these concepts are distinct, allowing a dataset to be characterized as accurate, precise, both, or neither.

Cohen's kappa statistic, employed in the evaluation process, serves as a measure of inter-rater reliability, synonymous with inter-observer agreement. Ranging from 0 to 1, the kappa statistic signifies different levels of agreement, from chance agreement (0) to perfect agreement (1).

(a) 0 = agreement equivalent to chance.
(b) 0.1–0.20 = slight agreement.
(c) 0.21–0.40 = fair agreement.
(d) 0.41–0.60 = moderate agreement.
(e) 0.61–0.80 = substantial agreement.
(f) 0.81–0.99 = near-perfect agreement.
(g) 1 = perfect agreement.

III.III Synthesizing Results

The final combined model, which integrates the strengths of various regression models, exhibits exceptional predictive performance for crop yield estimation. The final model is developed by averaging the values of the models using a cross-validation function, which runs cross validation on a dataset to test whether the model can generalize over the whole dataset. The function returns a list of scores per fold, and the average of these scores can be calculated to provide a single metric value for the dataset. The R-squared value of 0.940 ± 0.003 indicates that approximately 94% of the variance in crop yields can be explained by the model, showcasing its high explanatory power. This level of accuracy is particularly noteworthy in the context of agricultural yield prediction, where precise estimations are crucial for optimizing resource allocation and planning.

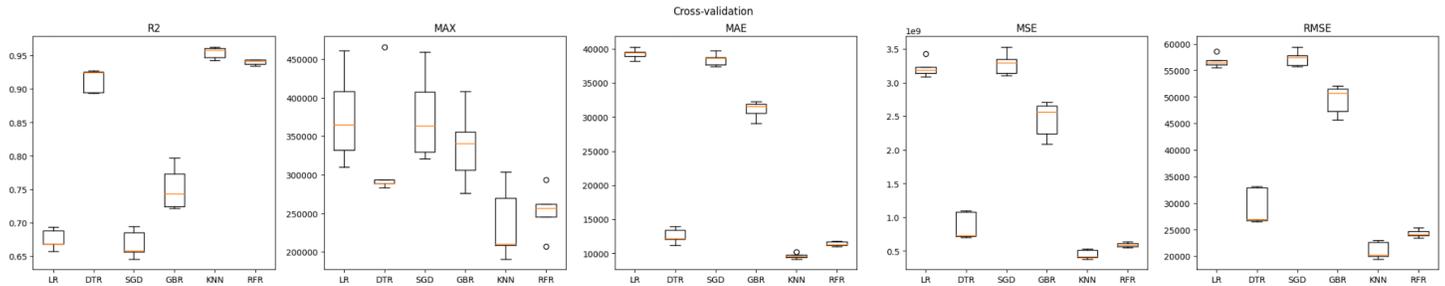

Figure 8: Synthesizing final combined models

The derivation of the final model involved combining outputs from various regression models. Analyzing feature importance or coefficients in the context of this ensemble model requires a nuanced approach. Techniques such as permutation importance or feature contribution from individual base models can offer insights into the most influential parameters affecting crop yields (Molnar et. al 1970).

$R^2$: 0.940 ± 0.003
MAX: 327,329 ± 101,418
MAE: 13,248 ± 229
MSE: 699,040,572 ± 38,623,281
RMSE: 26,429 ± 741

In the absence of specific feature importance details, the interpretation should consider the aggregated impact of temperature, rain, and insecticidal usage. Understanding how these variables collectively influence the ensemble model's predictions is crucial for practical implementation.

IV. Conclusion

IV.I Model Selection Rationale:

Our research meticulously employed six regression models—Linear, Tree, Gradient Descent, Gradient Boosting, K-Nearest Neighbors, and Random Forest—in a judicious selection process. We systematically summarized the performance of each model, scrutinizing their aptness for the given dataset. Notably, the Random Forest Regression model emerged, boasting a determination coefficient ($r^2$) of 0.95. This performance underscored its suitability for the complex task of predicting crop yields across 196 countries.

The appropriateness of each model was carefully considered in the context of the unique challenges posed by our dataset. The diverse nature of our training parameters—pesticides (tonnes), rainfall (mm), temperature (Celsius), and yield (hg/ha)—demanded models capable of capturing non-linear relationships and intricate dependencies. The rationale behind the model selection process was anchored in achieving both accuracy and interpretability, crucial factors in the nuanced domain of crop yield prediction.

IV.II Limitations and Future Work:

While our research presents significant advancements, it is imperative to acknowledge certain limitations. The study's reliance on historical data introduces a temporal constraint, limiting the model's ability to adapt to rapidly changing environmental conditions (Yu et. al 2018). Additionally, the interpretability challenges associated with ensemble models, like Random Forest, prompt avenues for further exploration into model explainability and transparency.

Future research endeavors should delve into refining the model's interpretability while preserving its exceptional predictive accuracy. Exploring additional factors, such as soil composition and pest prevalence, could enhance the model's comprehensiveness (Klompenburg 2020). Moreover, integrating real-time data sources and employing advanced optimization techniques may address limitations related to temporal constraints, ensuring the model's adaptability to dynamic agricultural landscapes (Araújo et. al 2021).

## AUTHORS

**First Author** – Ishaan Gupta, Engineering Department, Dublin High School, 8151 Village Parkway, Dublin CA, ishaangupta006@gmail.com

**Second Author** – Samyutha Ayalasomayajula, Engineering Department, Dublin High School, 8151 Village Parkway, Dublin CA, samyutha06@gmail.com

**Third Author** – Yashas Shashdhara, Engineering Department, Dublin High School, 8151 Village Parkway, Dublin CA, yashas2260@gmail.com

**Fourth Author** – Anish Kataria, Department of Computer Sciences, 35 Olden St, Princeton, NJ, ak8686@princeton.edu

**Fifth Author** – Shreyes Shashidara, Engineering Department, Dublin High School, 8151 Village Parkway, Dublin CA, shreyas.skys1@gmail.com

**Sixth Author** – Krishita Kataria, Engineering Department, Dublin High School, 8151 Village Parkway, Dublin CA, shreyas.skys1@gmail.com

**Guiding Author** – Aditya Undurti, aundurti@mit.edu